\documentclass{article}
\usepackage{spconf,amsmath,graphicx,hyperref}
\usepackage{booktabs}
\usepackage{amssymb}
\usepackage{xcolor}

\title{CauCLIP: Bridging the Sim-to-Real Gap in Surgical Video Understanding via Causality-Inspired Vision-Language Modeling}
\name{Yuxin He, An Li, Cheng Xue\textsuperscript{*} 
\thanks{This work was supported by the National Natural Science Foundation of China (grant no. 62401143), the State Key Project of Research and Development Plan (grants no. 2024YFF1206703), the Natural Science Foundation of Jiangsu Province (grant no. BK20241301).\\
\textsuperscript{*}Corresponding author: Cheng Xue}}
\address{ Southeast University, Nanjing, China \\
\{yuxinhe, lian, cxue\} @seu.edu.cn}

\begin{document}
%
\maketitle
\begin{abstract}
Surgical phase recognition is a critical component for context-aware decision support in intelligent operating rooms, yet training robust models is hindered by limited annotated clinical videos and large domain gaps between synthetic and real surgical data. To address this, we propose CauCLIP, a causality-inspired vision-language framework that leverages CLIP to learn domain-invariant representations for surgical phase recognition without access to target domain data. Our approach integrates a frequency-based augmentation strategy to perturb domain-specific attributes while preserving semantic structures, and a causal suppression loss that mitigates non-causal biases and reinforces causal surgical features.
These components are combined in a unified training framework that enables the model to focus on stable causal factors underlying surgical workflows. Experiments on the SurgVisDom hard adaptation benchmark demonstrate that our method substantially outperforms all competing approaches, highlighting the effectiveness of causality-guided vision-language models for domain-generalizable surgical video understanding.
\end{abstract}

%
\section{Introduction}
\label{sec:intro}
Surgical phase recognition is a fundamental building block toward context-aware decision support in intelligent operating rooms. Accurate recognition of surgical steps enables real-time feedback, autonomous robotic assistance, and retrospective workflow analysis. However, learning robust models for surgical phase recognition remains challenging due to the scarcity of large-scale, labeled clinical video datasets. The high cost of data annotation, concerns around patient privacy, and variability in surgical procedures limit the availability of real-world data for training deep learning models.

To address this, the MICCAI SurgVisDom challenge~\cite{surgvisdomdataset} introduced a benchmark for visual domain adaptation in surgical video understanding, providing data from two distinct domains: a virtual reality (VR) surgical simulator and real-world porcine surgery. Models trained solely on VR data often fail to generalize to clinical settings due to their shifts in visual characteristics: VR environments are typically bright, bloodless, and feature tools with clear outlines, whereas clinical surgery involves complex lighting, blood, smoke, and cluttered backgrounds. This sim-to-real problem requires a model that can generalize across domains without access to target domain data, a task formally known as domain generalization.
\begin{figure}[t]
    \centering
    \includegraphics[width=0.92\columnwidth]{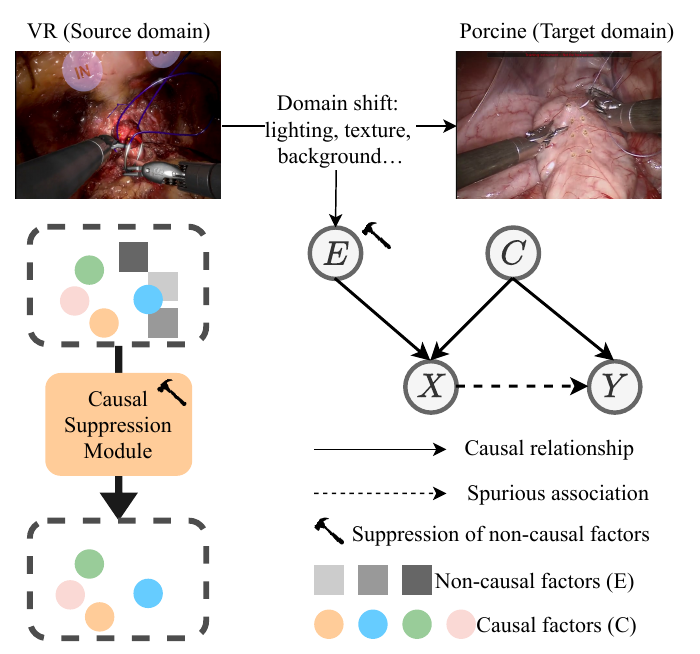}
    \vspace{-1mm}
    \caption{Illustration of the same task in different domains. Domain shift introduces non-causal factors (E) such as lighting and texture, while causal factors (C) remain task-relevant. Our causal suppression module reduces the influence of E and reinforces C.}
    \label{fig:f1}
    \vspace{-3mm}
\end{figure}

Mainstream domain generalization approaches include data augmentation and meta-learning. Although effective in moderate variations, these methods fundamentally rely on statistical correlations. Data augmentation increases data diversity through techniques like style mixing~\cite{adain}\cite{stylemix}\cite{poda}, yet it cannot model the evolving shifts in surgical videos, and still relies on superficial visual cues spuriously correlated with labels. Meta-learning~\cite{meta}\cite{meta1} simulates shifts via episodic domain splits, when the assumed splits mismatch the real shift process, the learned features may remain correlation-based and fail to generalize to unseen surgical conditions.

Vision-language models (VLMs) such as CLIP~\cite{clip} have shown strong transferability across vision tasks~\cite{clip4MI}, leveraging text supervision for generalization in low-data settings~\cite{sda}. However, their application in surgical phase recognition under domain shift remains largely unexplored, and naive application often relies on spurious domain-specific correlations. Causality provides a principled approach for domain generalization by emphasizing stable~\cite{CausalityDG}, task-relevant mechanisms while suppressing non-causal factors such as lighting, texture, and spurious instrument motions. Motivated by this insight, we develop a causality-inspired framework built upon CLIP, which suppresses domain-specific noise and strengthens domain-invariant surgical semantics to enhance generalization across domains.

Our approach integrates a frequency-based augmentation strategy that perturbs high-frequency, non-causal components while preserving semantic structures, along with a spurious suppression loss that enforces separation between domain-invariant and domain-specific features. These elements are combined in a unified training framework that enables multimodal supervision while mitigating reliance on spurious correlations. Experiments on the SurgVisDom hard adaptation benchmark demonstrate that our method significantly outperforms all competing approaches, highlighting the effectiveness of causality-guided VLMs for domain-generalizable surgical video understanding.


Our contributions are summarized as follows:
\vspace{-6pt}
\begin{itemize}
    \item We propose a causality-inspired vision-language framework for surgical phase recognition, which explicitly reinforces causal surgical semantics and attenuates domain-specific nuisances.
    \vspace{-6pt}
    \item We design a frequency-based augmentation strategy and an novel suppression loss that jointly attenuate spurious correlations (e.g., lighting, texture, and tool appearance) while enhancing invariance to domain shifts.  
    \vspace{-6pt}
    \item We develop a unified training framework that integrates these components into CLIP, enabling principled multimodal supervision and achieving state-of-the-art performance on the SurgVisDom hard adaptation benchmark, substantially outperforming all competing approaches.  
\end{itemize}
\vspace{-5pt}

\section{Method}
\label{sec:pagestyle}
\begin{figure*}[t]
    \centering
    \includegraphics[width=0.98\linewidth]{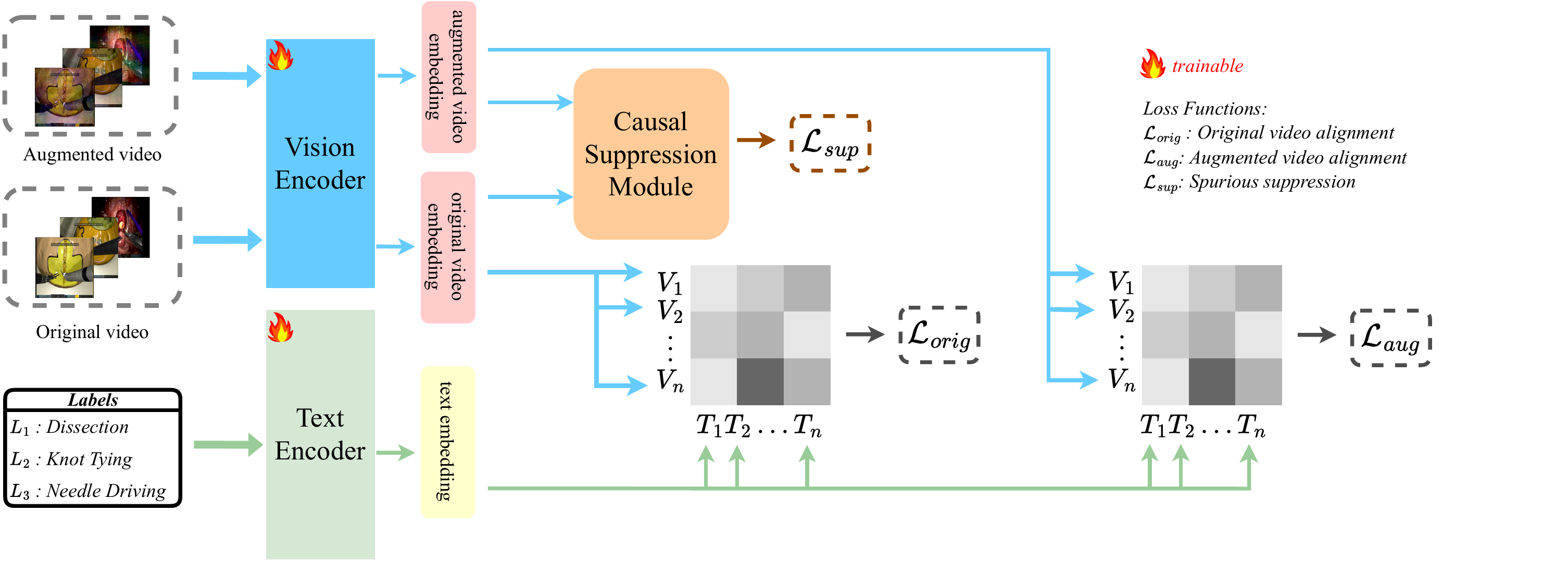}
    \vspace{-1mm}
    \caption{The framework of CauCLIP. The model learns visual-text alignment using original and augmented videos. A causal suppression module ensures that representations are invariant to stylistic perturbations introduced in the frequency domain.}
    \label{fig:model}
    \vspace{-3mm}
\end{figure*}
\subsection{CLIP-based Surgical Phase Recognition}
A schematic framework of our proposed model is shown in Fig~\ref{fig:model}. Our framework builds upon ActionCLIP~\cite{actionclip}, adapting it for surgical video understanding. We reformulate phase recognition as a video-text matching task~\cite{videoclip}\cite{clip4clip}, leveraging CLIP's pre-trained image-text alignment. ActionCLIP leverages the powerful image-text representation learning capabilities of CLIP and extends it to the video domain by incorporating a video encoder $f_v(\cdot)$ and a text encoder $f_t(\cdot)$. The video encoder (ViT backbone) extracts spatio-temporal features from the input video $V$ to produce a visual embedding $z_v = f_v(V)$. A text encoder $f_t(\cdot)$ embeds a set of predefined prompts $\{T_c\}_{c=1}^C$ into textual features $\{z_{t,c}\}_{c=1}^C$, where $C$ is the number of surgical classes.

The similarity between $z_v$ and $z_{t,c}$ is then calculated to perform action recognition, converted into a probability distribution using a softmax with a learnable temperature parameter $\tau$:
\begin{equation}
\begin{aligned}
    p_{v2t}(\mathbf{V})_i &= \frac{\exp(\operatorname{sim}(\mathbf{V}, T_i) / \tau)}{\sum_{j=1}^C \exp(\operatorname{sim}(\mathbf{V}, T_j) / \tau)} \\
    p_{t2v}(\mathbf{T})_i &= \frac{\exp(\operatorname{sim}(\mathbf{T}, V_i) / \tau)}{\sum_{j=1}^C \exp(\operatorname{sim}(\mathbf{T}, V_j) / \tau)}
\end{aligned}
.
\label{softmax_sim}
\end{equation}
 Following ActionCLIP, we compute video-to-text and text-to-video logits for each input clip. Let $y_{v2t}$ and $y_{t2v}$ represent the ground-truth similarity scores, where the negative pair has a possibility of 0 and the positive pair has a possibility of 1. The CLIP loss is defined as the average Kullback-Leibler (KL) divergence:
\begin{equation}
\mathcal{L}_{\text{CLIP}} = \frac{1}{2} \left[ D_{\text{KL}}(p_{v2t} , y_{v2t}) + D_{\text{KL}}(p_{t2v},y_{t2v}) \right],
\end{equation}
\vspace{-5pt}

\subsection{Causality-Inspired Suppression and Augmentation}
\label{subsec:causal}
Though VLMs have strong domain generalization capability, their visual encoder is prone to learning spurious correlation from source domain, such as textures or lighting.
From a causal perspective \cite{disentangle}\cite{VTdisentangled}, these domain-specific attributes are non-causal factors $E$, whereas the underlying motion and tool-tissue interactions are the true causal factors $C$ for recognizing the surgical task. 
Thus, we aim to suppress the influence of $E$ and encourage the model to focus on $C$, enforcing invariance to domain shifts.\\

\textbf{Frequency-Domain Causal Augmentation}
To counteract non-causal cues, we propose a frequency-based augmentation inspired by~\cite{CausalityDG}\cite{fourier} that perturbs amplitude while preserving phase, based on the observation that phase encodes high-level semantics, while amplitude reflects style.

A key challenge in our VR-only training  dataset is lack of explicit multiple “domains" for stylistic variation. However, we observe that significant stylistic variations exist in the backgrounds across different surgical tasks. We leverage this intra-domain diversity as a proxy for inter-domain style shifts. Specifically, given an image $x_o$, we extract amplitude $A(x_o)$ and phase $P(x_o)$ via Fourier transformation. We mix its amplitude with a second image $x'_o$ (sampled from a different background) and reconstruct an augmented view $x_a$:
\begin{equation}
    \mathcal{F}(x^o) = A(x^o) e^{-jP(x^o)},
\label{fourier}
\end{equation}

\begin{equation}
\widetilde{A}(x^a) = (1-\beta)A(x^o) + \beta A((x')^o),
\label{eq:amplitude_mix}
\end{equation}

\begin{equation}
x^a = \mathcal{F}^{-1}(\widetilde{A}(x^a)e^{-jP(x^o)}.
\label{eq:inverse_fourier}
\end{equation}

where $\beta$ is a mixing ratio sampled from a uniform distribution $U(0, \alpha)$, and $\alpha$ is a hyperparameter controlling the maximum degree of the augmentation. We found that setting \(\alpha = 0.5\) achieved optimal performance, balancing augmentation and the original semantic. The Fourier transformation $\mathcal{F}(\cdot)$ and its inverse $\mathcal{F}^{-1}(\cdot)$ can be calculated with the FFT algorithm.\\

\textbf{Causal Suppression Module}
We encourage the encoder to extract invariant features across $x_o$ and $x_a$ by enforcing similarity in their representations. Let $z_o = f_v(V_o)$ and $z_a = f_v(V_a)$ denote visual embeddings of the original and augmented clips.
We compute a N$\times$N cosine similarity matrix $C$ over a batch of size $N$, where each element $C_{ij}$ represents the cosine similarity between the $i$-th and $j$-th feature vectors in the combined set. We define the suppression loss as:
\begin{equation}
\mathcal{L}_{\text{sup}} = \|C - I\|_F^2,
\end{equation}
where $\| \cdot \|_F$ is the Frobenius norm, and $I$ is an identity matrix. It encourages high similarity between original and frequency-augmented views, while simultaneously enforcing decorrelation across different clips.
Our goal is to ensure corresponding original augmented pairs are highly correlated, while all other pairs (e.g., $\mathbf{z}^o_i$ and $\mathbf{z}^a_j$) are decorrelated to promote features independence.\\

\textbf{Augmented Alignment Loss.}  
We introduce an additional alignment loss $\mathcal{L}_{\text{aug}}$ to encourage the model to preserve semantic consistency across the augmented views, which is similar to the original CLIP loss. Frequency-domain augmentation perturbs low-level visual statistics, while preserving high-level semantics, enforcing alignment between augmented images and original textual prompt. It can discourage the model from exploiting spurious domain-specific cues and strengthens its reliance on invariant surgical semantics.\\

\textbf{Total Loss}
Our final loss function combines the original KL loss, the additional KL loss on augmented views $\mathcal{L}_{aug}$, and a causality-inspired suppression term $\mathcal{L}_{sup}$:
\begin{equation}
    \mathcal{L}_{total} = \mathcal{L}_{orig} + \lambda_{aug} \mathcal{L}_{aug} + \lambda_{sup} \mathcal{L}_{sup},
\end{equation}
where $\lambda_{aug}$ and $\lambda_{sup}$ are hyperparameters. 
\section{Experiment}
\label{sec:typestyle}

\subsection{Datasets}
We evaluate our model on the SurgVisDom dataset from the MICCAI 2020 EndoVis Challenge~\cite{surgvisdomdataset}. The dataset contains videos collected from both virtual reality (VR) and clinical-like porcine domains. It consists of 3 tasks: dissection (DS), knot-tying (KT) and needle-driving (ND).

The training set includes 450 VR and 27 clinical-like video clips, each containing a single task. The test set consists of 16 long clinical videos that mix multiple tasks. We follow the official protocol that only frames of valid phases are considered for evaluation. For hard domain adaptation in the challenge, only VR data is required to generalize real clinical-like data with severe domain shifts and class imbalance.
\vspace{-5pt}

\subsection{Evaluation Metrics}
\label{subsec:metrics}
We adopt the official evaluation metrics for the challenge, including weighted F1-score, unweighted F1-score, global F1-score and balanced accuracy. Weighted F1-score and the balanced accuracy explicitly accounts for class imbalance between phases. Unweighted F1-score and global F1-score, which correspond to macro F1 and micro F1 respectively, are used to test rank stability. 

\vspace{-5pt}

\subsection{Implementation Details}
\label{subsec: implement}
\begin{figure}[t!]
    \centering
    \includegraphics[width=0.85\linewidth]{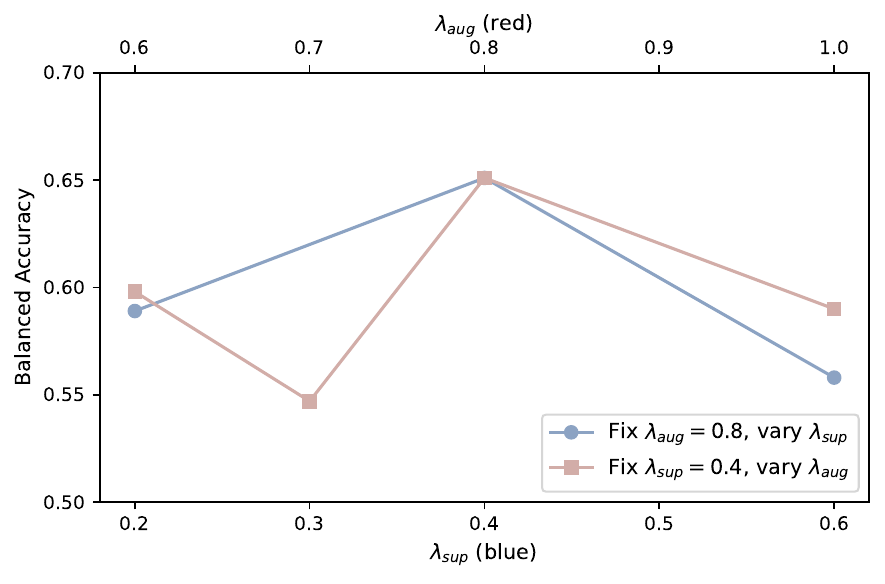}
    \vspace{-2mm}
    \caption{Analysis on $\lambda_{aug}$ and $\lambda_{sup}$ on balanced accuracy}
    \label{fig:line}
    \vspace{-3mm}
\end{figure}

We implement our model in PyTorch using a CLIP ViT-B/16 backbone. Both vision and text encoders are initialized from CLIP weights. Training is performed on NVIDIA RTX 4090 GPUs. During training, we uniformly sample sparse frames from each clip and resize them to $224 \times 224$ pixels. We employed multi-scale crop, random-horizontal flipping, color jittering and our proposed causality-inspired augmentation (Sec~\ref{subsec:causal}) for cross-domain robustness. We trained our model with AdamW optimizer, using a batch size of 32 and an initial learning rate of $1 \times 10^{-5}$. For the loss weight, we set $\lambda_{aug}$ = 0.8 and $\lambda_{sup}$ = 0.4 based on empirical validation (see Fig~\ref{fig:line}). For evaluation, we use the same frame sampling and standard augmentation pipeline. Predictions are aggregated by averaging logits across all sequences. 
\vspace{-5pt}
\begin{table}[t!]
    \centering
    \caption{Action Recognition Results of Different Models on Hard Domain Adaptation}
    \label{tab:results}
    \resizebox{\columnwidth}{!}{
    \begin{tabular}{@{}lcccc@{}}
    \toprule
    \textbf{Method} & \textbf{Weighted F1} & \textbf{Unweighted F1} & \textbf{Global F1} & \textbf{Balanced Accuracy} \\
    \midrule
    Rand~\cite{surgvisdomdataset} & 0.450 & 0.207 & 0.327 & 0.327 \\
    SK~\cite{surgvisdomdataset}   & 0.460 & 0.225 & 0.370 & 0.369 \\
Parakeet~\cite{surgvisdomdataset} & 0.470 & 0.266 & 0.475 & 0.559  \\
ResNet-50~\cite{resnet} & 0.497 & 0.355 & 0.583 & 0.425 \\ 
Vit-B/16~\cite{vit} & 0.518 & 0.388 & 0.564 & 0.524 \\
SDA-CLIP~\cite{sda} & 0.632 & 0.442 & 0.704 & 0.468 \\
    \midrule
 CauCLIP (ours)  & \textbf{0.642 }  & \textbf{0.487} & \textbf{0.708} & \textbf{0.651} \\
 \bottomrule
    \end{tabular}
    }
\vspace{-5pt}
\end{table}

\subsection{Experiment Result}
We compare CauCLIP with existing baselines as shown in Table~\ref{tab:results}. Earlier approaches rely on task-specific architectures, such as 2DVGG16~\cite{vgg} network + 3D convolution kernel (Team Parakeet) or SlowFast~\cite{slowfast} with instrument masks and contours (Team SK). While effective in constrained settings, these models struggle to generalize across domains due to their dependence on domain-specific features. We report results of random guessing following~\cite{surgvisdomdataset}. We further include other non-CLIP visual models, including ResNet-50~\cite{resnet} and ViT-B/16~\cite{vit},  under the same training protocol. In addition, we compare with a pioneer work SDA-CLIP~\cite{sda} by loading their released weights. Our model achieves superior performance across all evaluation metrics compared with prior approaches. The result highlights the benefit of integrating causal learning with vision-language alignment for stronger cross-domain generalization.
\vspace{-5pt}

\subsection{Ablation Study}
\label{subsec:ablation}
To investigate the contribution of each component in our method, we conduct an ablation study on the SurgVisDom dataset by progressively adding our proposed modules to the ActionCLIP baseline, which is only trained with the primary KL divergence loss ($\mathcal{L}_{orig}$). We evaluate the effect of the causality-inspired frequency-domain augmentation ($\mathcal{L}_{aug}$) and causal-inspired suppression loss ($\mathcal{L}_{sup}$), as shown in  Table~\ref{tab:ablation}.

The baseline achieves a balanced accuracy of 0.480, already competitive due to CLIP initialization. Adding $\mathcal{L}_{sup}$ improves to 0.544 by emphasizing causal factors, while $\mathcal{L}_{aug}$ boosts it to 0.533 by regularizing against diverse visual styles. Combining all components yields the best results.
This demonstrates that our proposed modules are complementary: augmentation improves robustness to style variations, while suppression strengthens causal feature dominance over spurious factors.
\vspace{-5pt}

\begin{table}[t!]
    \centering
    \caption{Ablation Study of the Loss Components on the SurgVisDom Hard Domain Adaptation Task}
    \label{tab:ablation}
    \resizebox{\columnwidth}{!}{%
    \begin{tabular}{@{}lcccc@{}}
    \toprule
    \textbf{Loss Function} & \textbf{Weighted F1} & \textbf{Unweighted F1} & \textbf{Global F1} & \textbf{Balanced Accuracy} \\
    \midrule
    $\boldsymbol{\mathcal{L}_{orig}}$ (without augmented data) & 0.401 & 0.305 & 0.396 & 0.480 \\
    $\boldsymbol{\mathcal{L}_{orig}} + \boldsymbol{\mathcal{L}_{sup}}$ & 0.524 & 0.394 & 0.562 & 0.544 \\
    $\boldsymbol{\mathcal{L}_{orig}} + \boldsymbol{\mathcal{L}_{aug}}$ & 0.540 & 0.402 & 0.591 & 0.533 \\
   \(\boldsymbol{\mathcal{L}_{orig}} + \boldsymbol{\mathcal{L}_{sup}} + \boldsymbol{\mathcal{L}_{aug}}\) (ours) & 0.642 & 0.487 & 0.708 & 0.651 \\
    \bottomrule
    \end{tabular}%
    }
\end{table}

\section{Conclusion}
\label{sec:majhead}

In this work, we present a causality-inspired CLIP framework for surgical phase recognition. By combining frequency-domain augmentation with a suppression module, our method emphasizes causal factors and reduces the reliance of non-causal cues, enabling robust sim-to-real transfer. Experiments on the SurgVisDom benchmark confirm that our approach achieves state-of-the-art performance. These results highlight the potential of integrating vision-language models with causal reasoning for reliable and generalizable surgical video understanding.
\vspace{-5pt}

\vspace{-1mm}

\bibliographystyle{IEEEbib}
\bibliography{strings,refs}

@inproceedings{clip,
  title={Learning transferable visual models from natural language supervision},
  author={Radford, Alec and Kim, Jong Wook and Hallacy, Chris and Ramesh, Aditya and Goh, Gabriel and Agarwal, Sandhini and Sastry, Girish and Askell, Amanda and Mishkin, Pamela and Clark, Jack and others},
  booktitle={International conference on machine learning},
  pages={8748--8763},
  year={2021},
  organization={PmLR}
}

@inproceedings{resnet,
  title={Deep residual learning for image recognition},
  author={He, Kaiming and Zhang, Xiangyu and Ren, Shaoqing and Sun, Jian},
  booktitle={Proceedings of the IEEE conference on computer vision and pattern recognition},
  pages={770--778},
  year={2016}
}

@article{vit,
  title={An image is worth 16x16 words: Transformers for image recognition at scale},
  author={Dosovitskiy, Alexey},
  journal={arXiv preprint arXiv:2010.11929},
  year={2020}
}

@inproceedings{slowfast,
  title={Slowfast networks for video recognition},
  author={Feichtenhofer, Christoph and Fan, Haoqi and Malik, Jitendra and He, Kaiming},
  booktitle={Proceedings of the IEEE/CVF international conference on computer vision},
  pages={6202--6211},
  year={2019}
}

@article{vgg,
  title={Very deep convolutional networks for large-scale image recognition},
  author={Simonyan, Karen and Zisserman, Andrew},
  journal={arXiv preprint arXiv:1409.1556},
  year={2014}
}

@article{surgvisdomdataset,
  title={Surgical visual domain adaptation: Results from the MICCAI 2020 SurgVisDom challenge},
  author={Zia, Aneeq and Bhattacharyya, Kiran and Liu, Xi and Wang, Ziheng and Kondo, Satoshi and Colleoni, Emanuele and Van Amsterdam, Beatrice and Hussain, Razeen and Hussain, Raabid and Maier-Hein, Lena and others},
  journal={arXiv preprint arXiv:2102.13644},
  year={2021}
}

@article{actionclip,
  title={Actionclip: A new paradigm for video action recognition},
  author={Wang, Mengmeng and Xing, Jiazheng and Liu, Yong},
  journal={arXiv preprint arXiv:2109.08472},
  year={2021}
}

@InProceedings{CausalityDG,
    author    = {Lv, Fangrui and Liang, Jian and Li, Shuang and Zang, Bin and Liu, Chi Harold and Wang, Ziteng and Liu, Di},
    title     = {Causality Inspired Representation Learning for Domain Generalization},
    booktitle = {Proceedings of the IEEE/CVF Conference on Computer Vision and Pattern Recognition (CVPR)},
    month     = {June},
    year      = {2022},
    pages     = {8046-8056}
}

@article{clip4clip,
  title={Clip4clip: An empirical study of clip for end to end video clip retrieval and captioning},
  author={Luo, Huaishao and Ji, Lei and Zhong, Ming and Chen, Yang and Lei, Wen and Duan, Nan and Li, Tianrui},
  journal={Neurocomputing},
  volume={508},
  pages={293--304},
  year={2022},
  publisher={Elsevier}
}

@article{videoclip,
  title={Videoclip: Contrastive pre-training for zero-shot video-text understanding},
  author={Xu, Hu and Ghosh, Gargi and Huang, Po-Yao and Okhonko, Dmytro and Aghajanyan, Armen and Metze, Florian and Zettlemoyer, Luke and Feichtenhofer, Christoph},
  journal={arXiv preprint arXiv:2109.14084},
  year={2021}
}

@article{clip4MI,
  title={Clip in medical imaging: A comprehensive survey},
  author={Zhao, Zihao and Liu, Yuxiao and Wu, Han and Wang, Mei and Li, Yonghao and Wang, Sheng and Teng, Lin and Liu, Disheng and Cui, Zhiming and Wang, Qian and others},
  journal={arXiv preprint arXiv:2312.07353},
  year={2023}
}

@inproceedings{poda,
  title={Poda: Prompt-driven zero-shot domain adaptation},
  author={Fahes, Mohammad and Vu, Tuan-Hung and Bursuc, Andrei and P{\'e}rez, Patrick and De Charette, Raoul},
  booktitle={Proceedings of the IEEE/CVF International Conference on Computer Vision},
  pages={18623--18633},
  year={2023}
}

@InProceedings{meta,
    author    = {Shu, Yang and Cao, Zhangjie and Wang, Chenyu and Wang, Jianmin and Long, Mingsheng},
    title     = {Open Domain Generalization with Domain-Augmented Meta-Learning},
    booktitle = {Proceedings of the IEEE/CVF Conference on Computer Vision and Pattern Recognition (CVPR)},
    month     = {June},
    year      = {2021},
    pages     = {9624-9633}
}

@article{sda,
  title={SDA-CLIP: Surgical visual domain adaptation using video and text labels},
  author={Li, Yuchong and Jia, Shuangfu and Song, Guangbi and Wang, Ping and Jia, Fucang},
  journal={Quantitative Imaging in Medicine and Surgery},
  volume={13},
  number={10},
  pages={6989},
  year={2023}
}

@article{stylemix,
  title={Domain generalization with mixstyle},
  author={Zhou, Kaiyang and Yang, Yongxin and Qiao, Yu and Xiang, Tao},
  journal={arXiv preprint arXiv:2104.02008},
  year={2021}
}

@inproceedings{adain,
  title={Arbitrary style transfer in real-time with adaptive instance normalization},
  author={Huang, Xun and Belongie, Serge},
  booktitle={Proceedings of the IEEE international conference on computer vision},
  pages={1501--1510},
  year={2017}
}

@inproceedings{meta1,
  title={Generalizable decision boundaries: Dualistic meta-learning for open set domain generalization},
  author={Wang, Xiran and Zhang, Jian and Qi, Lei and Shi, Yinghuan},
  booktitle={Proceedings of the IEEE/CVF International Conference on Computer Vision},
  pages={11564--11573},
  year={2023}
}

@article{VTdisentangled,
  title={Disentangled representation learning for text-video retrieval},
  author={Wang, Qiang and Zhang, Yanhao and Zheng, Yun and Pan, Pan and Hua, Xian-Sheng},
  journal={arXiv preprint arXiv:2203.07111},
  year={2022}
}

@article{disentangle,
  title={Learning causally disentangled representations via the principle of independent causal mechanisms},
  author={Komanduri, Aneesh and Wu, Yongkai and Chen, Feng and Wu, Xintao},
  journal={arXiv preprint arXiv:2306.01213},
  year={2023}
}

@article{fourier,
  title={Causality-Based Contrastive Incremental Learning Framework for Domain Generalization},
  author={Wang, Xin and Zhao, Qingjie and Wang, Lei and Liu, Wangwang},
  journal={Tsinghua Science and Technology},
  volume={30},
  number={4},
  pages={1636--1647},
  year={2025},
  publisher={TUP}
}

\end{document}